\documentclass[11pt]{article}
\usepackage{comment}
\usepackage{acl}
\usepackage{times}
\usepackage{latexsym}
\usepackage[T1]{fontenc}
\usepackage[utf8]{inputenc}
\usepackage{microtype}
\usepackage{inconsolata}
\usepackage{graphicx}

\usepackage{bm} % Add this to your preamble
\usepackage{multirow}
\usepackage{array}
\usepackage{algorithm}
\usepackage{algpseudocode}
\usepackage{amsmath}
\usepackage{amssymb}
\usepackage{tabularx}
\usepackage{makecell}
\usepackage{threeparttable}
\usepackage{balance}
\usepackage{placeins}
\usepackage{booktabs}
\usepackage{pifont}

\usepackage{url}
\title{AgenticAI-DialogGen: Topic-Guided Conversation Generation for Fine-Tuning and Evaluating Short- and Long-Term Memories of LLMs}

\author{
Manoj Madushanka Perera, Adnan Mahmood, Kasun Eranda Wijethilake, and Quan Z. Sheng \\
School of Computing, Macquarie University, Sydney, NSW 2109, Australia \\
\texttt{manojmadushanka.perera@hdr.mq.edu.au}, \texttt{adnan.mahmood@mq.edu.au}, \\
\texttt{kasuneranda.wijethilake@hdr.mq.edu.au}, \texttt{michael.sheng@mq.edu.au}
}

\begin{document}
\maketitle
\begin{abstract}

Recent advancements in Large Language Models (LLMs) have improved their ability to process extended conversational contexts, yet fine-tuning and evaluating short- and long-term memories remain difficult due to the absence of datasets that encode both short- and long-term conversational history. Existing conversational datasets lack memory grounding, overlook topic continuity, or rely on costly human annotation. To address these gaps, we introduce AgenticAI-DialogGen, a modular agent-based framework that generates persona-grounded and topic-guided conversations without human supervision. The framework uses LLM agents to extract knowledge graphs, identify topics, build speaker personas, and simulate topic-guided conversations from unstructured conversations. A QA module generates memory-grounded Question Answer (QA) pairs drawn from short- and long-term conversational histories. We also generated a new dataset entitled, TopicGuidedChat (TGC), where long-term memory is encoded as speaker-specific knowledge graphs and short-term memory as newly generated topic-guided conversations. Evaluations depict that AgenticAI-DialogGen yields higher conversational quality and LLMs fine-tuned on TGC dataset achieve improved performance on memory-grounded QA tasks.

\end{abstract}

%\keywords{Large Language Models, Conversational AI, Agentic-AI, Conversational Dataset, Memory Modeling.}

\section{Introduction}

Advances in the context length and memory capacity of Large Language Models (LLMs) have improved their ability to handle extended conversations in conversational AI systems \cite{LLMcontext,LLMsContext, MMPT1,MMPT2}. With greater context retention, modern LLMs support conversations that span multiple topics, evolve over time, and reflect speaker traits \cite{MultiTurnNew,Topi}. However, longer context windows alone do not ensure reliable recall or consistent preservation of speaker traits and topic continuity. Achieving these abilities requires fine-tuning and evaluation on datasets that encode both short- and long-term memories, yet existing datasets lack such structure, limiting progress in memory-grounded conversation modeling.

\begin{figure}[!t]
  \centering
  \includegraphics[width=\linewidth]{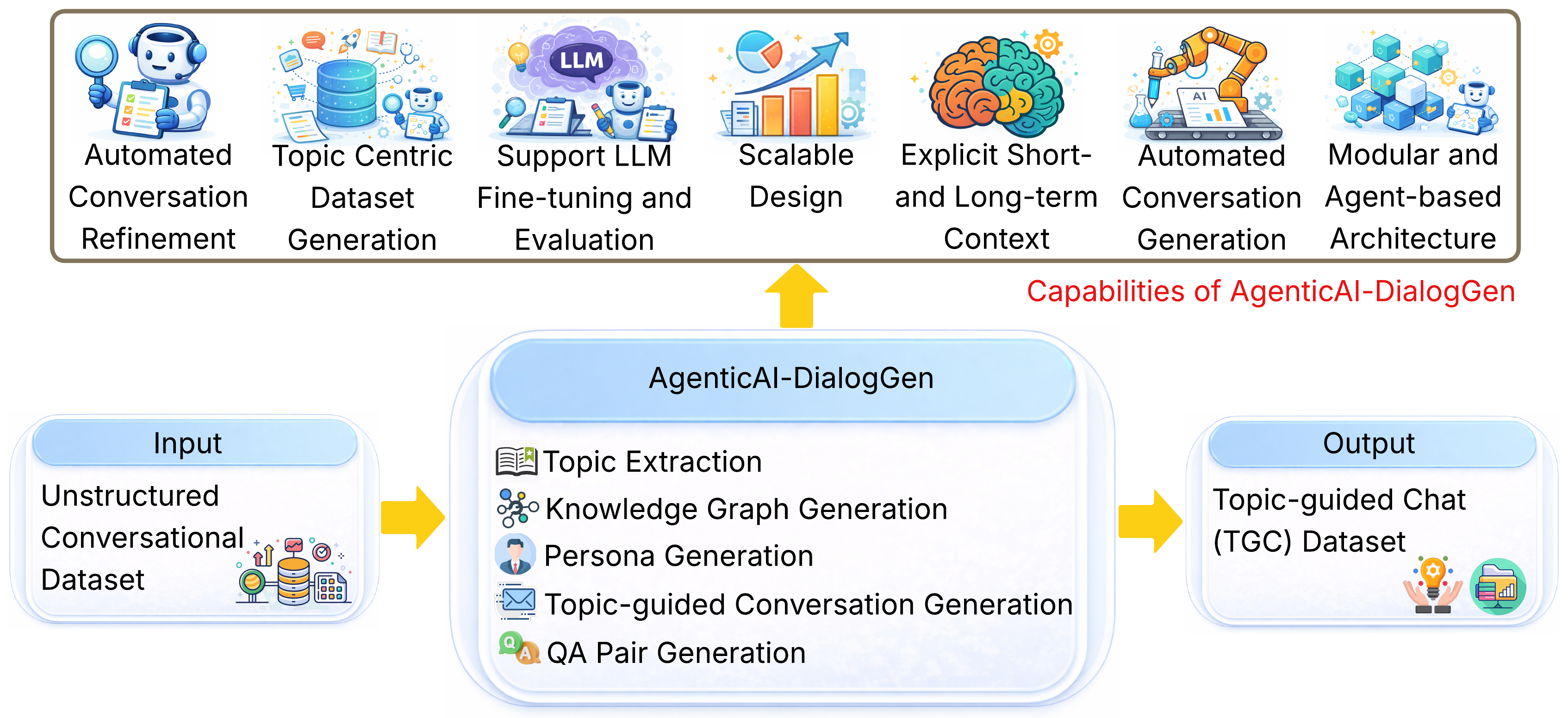}
  \vspace{-1em}
  \caption{A high-level overview of the AgenticAI-DialogGen framework.}
  \label{fig:framework2}
  \vspace{-5mm}
\end{figure}

Existing conversational datasets fall into two main categories. The first category includes conversational question answering datasets, i.e., CoQA \cite{coqa}, QuAC \cite{quac}, TopiOCQA \cite{TopiOCQA}, and SQuAD 2.0 \cite{SQAUD}, which evaluate multi-turn Question Answering (QA) over short contexts \cite{ConvQA}. These datasets 
evaluate coherence and co-reference resolution but lack persona grounding and topic continuity. The second category includes open-domain conversational datasets, i.e., PersonaChat \cite{PersonaChat}, DailyDialog \cite{dailydialog}, EmpatheticDialogues \cite{Empathetic}, and Multi Session Chat (MSC) \cite{MSCDataset}, which capture interpersonal dynamics, speaker traits, and free-form conversations. However, they lack QA behavior, structured coherence, and memory grounded tasks. Amongst them, MSC offers long-context, multi-session conversations yet lacks topic structure and organizes conversations by sessions instead of coherent topics.

Despite the advances in multi-turn conversation modeling, existing conversational datasets remain insufficient for fine-tuning and evaluating LLMs on memory recall and topic continuity. These capabilities require conversational datasets that support realistic, persona-consistent, and topic-guided conversations with both short- and long-term memories. Creating such datasets manually is costly and error-prone \cite{MannualAnno} as human annotators struggle to maintain coherence, preserve persona traits, and represent extended context. These limitations highlight the need for automated frameworks that generate high-quality, topic-guided, and persona-grounded conversations at scale without human supervision.

To address these limitations, we introduce AgenticAI-DialogGen, a modular agent-based framework for generating high-quality, persona-grounded, and topic-guided conversations. The framework employs LLM agents to extract knowledge, identify topics, generate personas, simulate multi-turn conversations, and refine conversations without human supervision. It transforms unstructured conversations into structured knowledge graphs and simulates conversations using LangGraph-based agents guided by speaker traits and topical interests. All conversations are validated and refined for topical adherence and quality. To support short- and long-term memory evaluation, the framework also generates QA pairs grounded in speaker-specific knowledge graphs and simulated conversational turns. AgenticAI-DialogGen is developed and evaluated primarily on 
the MSC dataset and further tested on the 
PersonaChat and DailyDialog datasets to confirm robustness across domains.

Using AgenticAI-DialogGen, we introduce the TopicGuidedChat (TGC) dataset, a synthetic benchmark derived from the MSC dataset. Unlike the session-based MSC dataset, TGC organizes conversations around semantically coherent topics and encodes both short- and long-term memories for each speaker pair. Long-term memory consists of speaker-specific knowledge graphs built from MSC dataset while short-term memory consists of newly generated topic-guided conversational turns. Furthermore, this design enables fine-tuning and evaluation using either structured knowledge graphs or unstructured MSC conversations as long-term memory. Moreover, TGC includes memory-grounded QA pairs for probing each speaker’s conversational history which allows LLMs to evaluate their short- and long-term memory retention. Figure~\ref{fig:framework2} presents a high-level overview of AgenticAI-DialogGen, highlighting its core components and design goals for topic-guided conversational dataset generation.

We assess how combining knowledge graphs with topic-guided conversations improves memory-aware behavior in LLMs. The results depict that AgenticAI-DialogGen enhances conversation quality and topic coherence over baselines. Furthermore, lightweight LLMs fine-tuned on the TGC dataset outperform larger zero-shot models, thereby demonstrating TGC’s practical utility for memory-grounded tasks. This work makes the following key contributions:

\begin{enumerate}
\item We present AgenticAI-DialogGen, i.e., a modular agent-based framework, that generates persona-grounded and topic-guided conversations without human supervision.
\item We release the TGC dataset for offering short-term context and long-term context in two forms: token-efficient knowledge graphs as structured memory and MSC conversations as unstructured memory, thereby making it suitable for diverse memory-grounded tasks.
\item We demonstrate through comprehensive evaluation that AgenticAI-DialogGen and TGC dataset enhance conversation quality, support memory-grounded QA tasks, and enable lightweight LLMs to demonstrate strong memory-aware behavior.
\end{enumerate}

\section{Related Work}\label{2}

Conversational question answering and open-domain conversational datasets have advanced multi-turn conversational AI systems. CoQA \cite{coqa} and QuAC \cite{quac} introduced multi-turn QA tasks where each question depends on prior conversational context while SQuAD 2.0 \cite{SQAUD} incorporated unanswerable questions to assess model uncertainty. TopiOCQA \cite{TopiOCQA} further refined conversational question answering by emphasizing topic continuity across related question sequences. Despite their contributions, these datasets remain constrained by manual annotation and fixed passage settings, restricting coverage, and context diversity. Open-domain conversational datasets, i.e., PersonaChat \cite{PersonaChat} and DailyDialog \cite{dailydialog}, support persona-grounded and free-form conversations but suffer from short length, limited topical range and QA tasks. The Multi-Session Chat (MSC) dataset \cite{MSCDataset} extended this by including session-wise conversations between recurring speaker pairs, yet lacks topic structure and remains costly to annotate. 

Recent work increasingly leveraged LLMs to automate conversational dataset generation with limited human supervision. SpeechDialogueFactory \cite{SpeechDialogueFactory} and ConvRecStudio \cite{chhetri2025convrecstudio} proposed scalable LLM-based frameworks for synthetic speech and conversational recommendation data generation, grounded in structured simulation and real user--item interactions. Furthermore, ConvoGen \cite{ConvoGen} adopted a multi-agent framework for conversation generation. Moreover, a modular system \cite{AnLLMbased} combined templates with user logic for multi-domain datasets while DialogueForge \cite{DialogueForge} enabled controllable chatbot interaction simulation. Despite improved scalability, these approaches did not focus topic-centric structure or persistent speaker-level memory across short- and long-term conversations. In contrast, AgenticAI-DialogGen integrates speaker-specific knowledge graphs with topic-guided conversations to evaluate short- and long-term memories of LLMs.

\section{Methodology}

\begin{figure}[!t]
  \centering
  \includegraphics[width=1\linewidth]{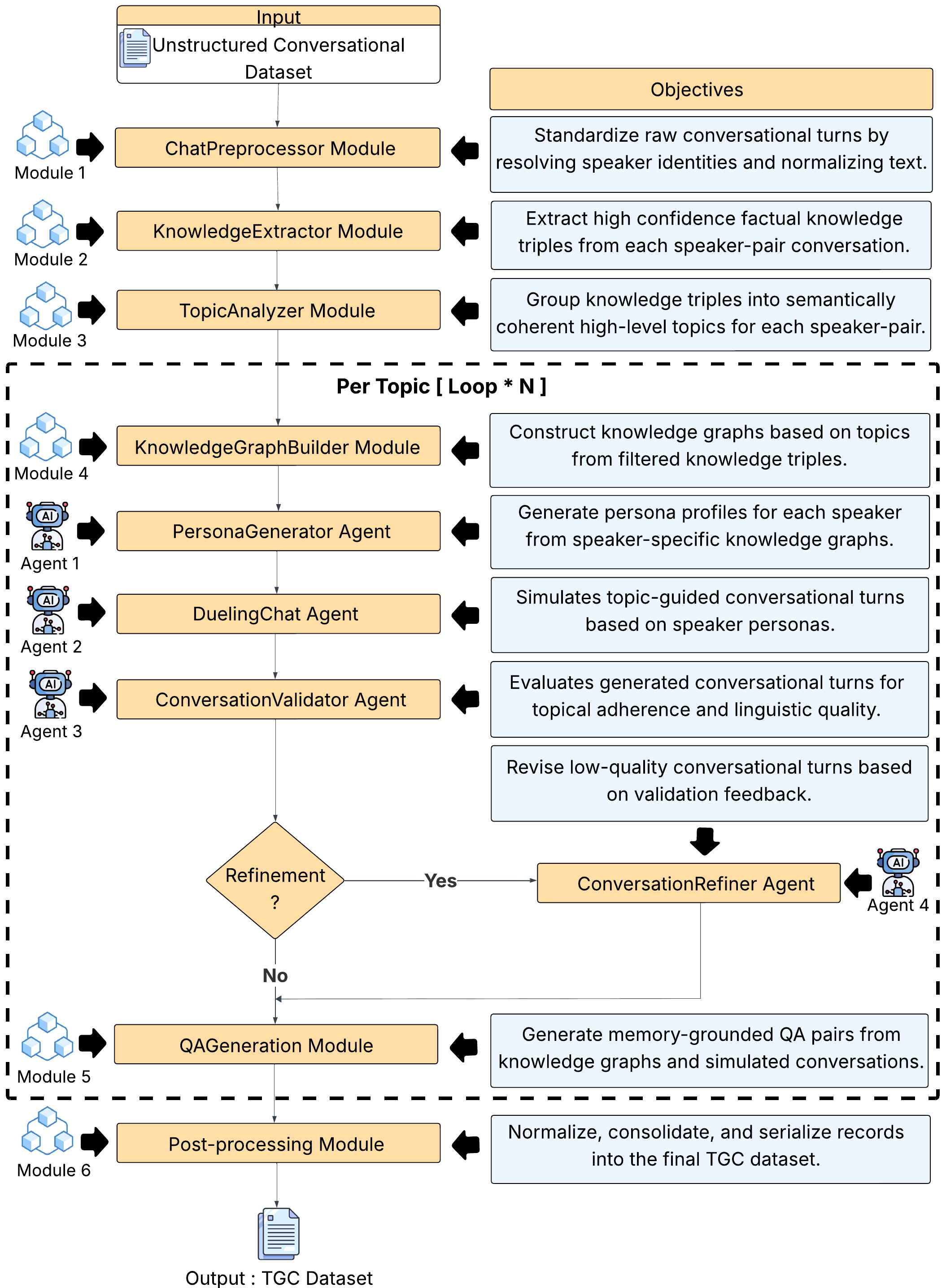}
  \vspace{-1em} 
  \caption{A modular agent-based architecture of AgenticAI-DialogGen.}
  \label{fig:framework}
  \vspace{-5mm}
\end{figure}

AgenticAI-DialogGen is a modular agent-based framework for generating high-quality, persona-grounded, and topic-guided conversations as delineated in Figure \ref{fig:framework}. Unlike approaches reliant on human annotation, it uses LLM agents to simulate conversations aligned with speaker traits and topical interests. The framework accepts any unstructured conversational dataset as input. In this work, the MSC dataset is used to construct the TGC dataset. We include only speaker pairs with four sessions (1,001 pairs), ensuring sufficient conversational depth for topic extraction, knowledge graph construction, and persona modeling. The framework is designed with transparency, modularity, and error tolerance, allowing each component to operate independently with clear input–output boundaries. This architecture supports scalable generation of topic-guided conversations while preserving persona coherence and topical alignment. %Implementation and configuration details are deferred to the supplementary material.

\subsection{Preprocessing Conversations}

The AgenticAI-DialogGen framework begins by preprocessing the conversational history for each speaker pair in a given source conversational dataset. The objective is to transform unstructured conversational turns in the source dataset into a clean and speaker-resolved sequence suitable for structured analysis. The \textit{ChatPreprocessor} module loads the source dataset conversational turns, assigns consistent speaker identifiers (\textit{speaker1} and \textit{speaker2}), and resolves ambiguities in the original text. An LLM-based prompt replaces first- and second-person pronouns with the correct speaker identifiers, followed by light text cleaning and the removal of incomplete conversational turns. The output is an ordered sequence of speaker-resolved conversational turns $\mathcal{T}$ indexed by turn position, defined as:

{\small
\begin{equation}
\mathcal{T} = \left(n(u_t), x'_t, t\right)_{t=1}^{T_{source}}
\end{equation}
}

\noindent where, $u_t$ is the original speaker id for turn $t$, $n(u_t)$ is the assigned speaker identifier in \{\textit{speaker1}, \textit{speaker2}\}, $x'_t$ is the refined text, and $t$ is the chronological turn index. The collection $\mathcal{T}$ represents an ordered sequence indexed by $t$, with $T_{source}$ denoting the total number of conversational turns for this speaker pair in the source dataset. This step ensures that all conversational turns are explicit, consistent, and ready for downstream modules.

\subsection{Knowledge Triple Extraction}

Following preprocessing, the complete conversation is passed to the \textit{KnowledgeExtractor} module which converts the rewritten conversational turns into structured factual knowledge. The module processes the entire conversation jointly, enabling the LLM to capture direct statements, cross-turn links, and implicit relationships. The processed conversational turns are formatted into a conversation-level prompt, and the LLM extracts subject--relation--object triples together with their corresponding source turn indices. Given the preprocessed conversational sequence $\mathcal{T}$, the \textit{KnowledgeExtractor} module produces a set of triple collections 
$\{\mathcal{K}_t\}_{t=1}^{T_{source}}$
where, each $\mathcal{K}_t$ contains the triples (subject (\(s\)), relation (\(r\)), and object (\(o\))) grounded in turn $t$. The extracted triples are validated, malformed triples are discarded, and the remaining triples are filtered by confidence score, de-duplicated, and attached to their respective turns. The final output is the union of all valid triples, defined as:

{\small
\begin{equation}
\mathcal{K} = \bigcup_{t=1}^{T_{source}} \mathcal{K}_t
\end{equation}
}

Each extracted triple retains a reference to its originating turn index $t$, enabling traceability across subsequent modules. This step converts free-form conversational text into a structured set of factual relationships that supports topic extraction, persona and knowledge graph generation.

\subsection{Topic Extraction}

After extracting structured knowledge triples, the next phase identifies core topics that organize the conversations. The \textit{TopicAnalyzer} module uses a prompt-guided LLM to group semantically related triples into coherent topic clusters, thereby uncovering dominant areas of shared interest that guide persona generation and conversation simulation. The module receives the complete set of valid triples \(\mathcal{K}\) and returns a set of topic groups as:

{\small
\begin{equation}
{T}_{\text{topics}} = \{t_j = (\text{name}_j, w_j, \alpha_j)\}_{j=1}^{J}
\end{equation}
}

\noindent where, \(J\) is the number of extracted topics, \(\text{name}_j\) is a descriptive label, \(w_j\) is its keyword set, and \(\alpha_j \in [0,1]\) is the LLM-assigned importance score. For each topic \(t_j\), the topic-specific triples associated with each speaker identifier \(n\) are:

{\small
\begin{equation}
\mathcal{K}_j^{n} =
\left\{
(s, r, o) \in \mathcal{K}\ \middle|\ 
\begin{aligned}
&(s, r, o) \text{ is assigned to topic } t_j \\
&\text{and } n \in \{\textit{speaker1}, \textit{speaker2}\}
\end{aligned}
\right\}
\end{equation}
}

A topic is retained only if both speakers appear in its assigned triples. The top \(N\) topics, ranked by \(\alpha_j\), are selected for downstream simulation. The final output is a ranked list of up to \(N\) \textit{TopicGroup} objects per speaker pair, each containing a topic name, keywords, topic-specific triples, involved speakers, and an importance score.

\subsection{Construction of Knowledge Graphs}

To structure the extracted knowledge and support persona modeling, the \textit{KnowledgeGraphBuilder} module converts topic-level triples into formal knowledge graphs. For each selected topic $t_j$, the associated topic-level triple set, obtained by aggregating speaker-specific triples, is transformed into a directed multigraph as:

{\small
\begin{equation}
\mathcal{G}_j = (V_j, E_j)
\end{equation}
}

\noindent where, $V_j$ contains all subjects and objects, and $E_j$ contains labeled edges representing semantic relations. The graphs are implemented using \textit{networkx.MultiDiGraph} to allow multiple relation types between the same node pairs. After constructing $\mathcal{G}_j$, the triples are filtered to obtain speaker-specific subsets. For each speaker identifier $n \in \{\textit{speaker1}, \textit{speaker2}\}$, the triples attributed to $n$ are retained to form the personalized knowledge set $\mathcal{K}_j^{n}$. This isolates the facts relevant to each speaker and provides the structured foundation for persona generation and subsequent conversation simulation.

\subsection{Persona Generation for Each Speaker}

With topic-specific knowledge graphs constructed for each speaker, the framework generates a personalized persona profile. The \textit{PersonaGenerator} agent processes the speaker-specific knowledge sets $\mathcal{K}_j^{n}$ for a given topic \(t_j\) and infers personality traits and topic interests that characterize the speaker’s conversational behavior. A structured persona profile \(\mathcal{P}_{n}\) is defined as:

{\small
\begin{equation}
    \mathcal{P}_{n} = \left( n,\ \mathcal{K}_j^{n},\ \{pt_1, pt_2, \ldots\},\ \{z_1, z_2, \ldots\} \right)
\end{equation}
}

\noindent where, \(n\) is the speaker identifier, \(\mathcal{K}_j^{n}\) is the associated speaker-specific knowledge triples, \(pt_i\) are inferred personality traits, and \(z_i\) are topic interests. These persona profiles guide downstream conversation generation by encoding the semantic and stylistic cues to simulate context-aware and personality-consistent conversations.

\subsection{Simulation of Topic-guided Conversations}

Using the persona profiles defined in the previous step, the framework simulates topic-guided conversations with the \textit{DuelingChatAgent} module. The simulation is implemented using the LangGraph framework that enforces turn alternation, shared memory, and controlled conversational flow. For distinct speaker identifiers \(n_1, n_2 \in \{\textit{speaker1}, \textit{speaker2}\}\), the agent \(A_{n_1}\) generates responses on odd-numbered turns and \(A_{n_2}\) on even-numbered turns. Each turn conditions on the evolving history buffer \(\mathcal{H}_t\). The conversation is generated for a fixed number of turns \(T\). The resulting conversation for topic \(t_j\) is defined as:

{\small
\begin{equation}
\mathcal{C}_j^{\text{sim}}
= \left\{ (s_t, x_t) = A_{n_t}(\mathcal{H}_t) \right\}_{t=1}^{T},
\quad
n_t =
\begin{cases}
n_1, & t\ \text{odd} \\
n_2, & t\ \text{even}
\end{cases}
\end{equation}
}

\noindent where, $s_t$ and $x_t$ denote the active speaker and generated text at turn $t$. The history buffer $\mathcal{H}_t = \{(s_k, x_k)\}_{k=1}^{t-1}$ contains all previously generated conversational turns up to turn $t-1$. This module produces fluent, persona-consistent, and topic-grounded conversations suitable for memory evaluation.

\subsection{Validation of Generated Conversations}

After simulation, each conversation undergoes automatic quality assessment using the \textit{ConversationValidator} agent. The agent receives the simulated conversation \( \mathcal{C}_j^{\text{sim}} \), the associated topic \( t_j \), and the speaker-specific knowledge sets \(\mathcal{K}_j^{n}\), and produces a structured validation object as:

{\small
\begin{align}
\mathcal{VA}_j = (&\texttt{is\_on\_topic},\ \texttt{adherence\_score}, \notag \\
                &\texttt{quality\_score},\ \texttt{issues},\ \texttt{suggestions})
\end{align}
}

\noindent where, \textit{is\_on\_topic} is a binary flag, \textit{adherence\_score} and \textit{quality\_score} take values from [0,1], and \textit{issues} and \textit{suggestions} list detected problems and recommended improvements. This phase identifies issues, i.e., off-topic drift, repetitive phrasing, or weak engagement. This validation provides a control signal for filtering low-quality conversations and supports the subsequent refinement stage, ensuring that only coherent and topic-consistent conversational turns are preserved.

\subsection{Refinement of Generated Conversations}

If a generated conversation \( \mathcal{C}_j^{\text{sim}} \) does not meet the minimum topical or quality threshold, the framework triggers a refinement step using validation feedback \( \mathcal{VA}_j \). The \textit{ConversationRefiner} agent prompts an LLM to revise the conversation while keeping the fixed turn count \(T\) and the strict alternation between \textit{speaker1} and \textit{speaker2}. The agent improves phrasing, coherence, and topical focus based on issues listed in \( \mathcal{VA}_j \). If the refined output is malformed, e.g., missing turns or not parseable, the system falls back to the original conversation. The final output of this module is:

{\small
\begin{equation}
\mathcal{C}_j^{\text{final}} =
\begin{cases}
\mathcal{C}_j^{\text{refined}}, & \texttt{if structurally valid} \\
\mathcal{C}_j^{\text{sim}}, & \texttt{otherwise}
\end{cases}
\end{equation}
}

This ensures only structurally and semantically consistent conversations enter the final dataset.

\subsection{Memory Grounded QA Generation}

To evaluate short- and long-term memories, the framework generates QA pairs grounded in speaker knowledge graphs and the simulated conversations. It simulates memory probing, where \textit{speaker1} asks about a prior event or preference and \textit{speaker2} provides a factual reply. The \textit{QAGeneration} module draws from two sources: (1) the speaker-specific knowledge sets \(\mathcal{K}_j^{n}\) (long-term memory) and (2) the final simulated conversation \(\mathcal{C}_j^{\text{final}}\) (short-term memory). The module produces a natural question from \textit{speaker1} and an answer from \textit{speaker2} depicted as: %Valid QA pairs for each topic \(t_j\) are stored as follows.

{\small
\begin{equation}
\mathcal{QA}_j = \left\{ (q_i, a_i) \right\}_{i=1}^{N_j}
\end{equation}
}

\noindent where, \(N_j\) is the number of valid QA pairs for topic \(t_j\). Generation continues until $N_j = n_{\text{target}}$ is reached. Candidate pairs are checked for duplication, correctness, and formatting, with invalid pairs removed. This module supplies retrieval-style supervision for assessing memory-grounded tasks across topic-guided and persona-grounded conversations.

The modules from \textit{KnowledgeGraphBuilder} to \textit{QAGeneration} repeat for each of the top \(N\) selected topics \(t_j\).

\renewcommand{\arraystretch}{1}
\begin{table*}[!t]
\footnotesize
\centering
\captionsetup{justification=centering}
\caption{Comparison of the TGC dataset with existing conversational datasets.}
\vspace{-2mm}
\label{tab:dataset_comparison}
\begin{tabular}{
    >{\centering\arraybackslash}m{5.0cm}|
    >{\centering\arraybackslash}m{0.7cm}|
    >{\centering\arraybackslash}m{0.7cm}|
    >{\centering\arraybackslash}m{0.7cm}|
    >{\centering\arraybackslash}m{0.7cm}|
    >{\centering\arraybackslash}m{0.7cm}|
    >{\centering\arraybackslash}m{0.7cm}|
    >{\centering\arraybackslash}m{0.7cm}|
    >{\centering\arraybackslash}m{0.7cm}|
    >{\centering\arraybackslash}m{0.7cm}
}
\hline\hline
\textbf{Dataset} & \textbf{MT} & \textbf{OD} & \textbf{PG} & \textbf{TC} & \textbf{KG} & \textbf{LT} & \textbf{ST} & \textbf{QA} & \textbf{AG} \\
\hline
\textbf{TGC (Ours)}        & \textbf{\boldmath$\checkmark$} & \textbf{\boldmath$\checkmark$} & \textbf{\boldmath$\checkmark$} & \textbf{\boldmath$\checkmark$} & \textbf{\boldmath$\checkmark$} & \textbf{\boldmath$\checkmark$} & \textbf{\boldmath$\checkmark$} & \textbf{\boldmath$\checkmark$} & \textbf{\boldmath$\checkmark$} \\
MSC \cite{MSCDataset}                & \checkmark & \checkmark & \checkmark & x          & x          & \checkmark & x          & x          & x \\
TopiOCQA \cite{TopiOCQA}             & \checkmark & \checkmark & x          & \checkmark & x          & x          & x          & \checkmark          & x \\
CoQA \cite{coqa}                     & \checkmark & x          & x          & x & x          & x          & x & \checkmark & x \\
PersonaChat \cite{PersonaChat}       & \checkmark & \checkmark & \checkmark & x          & x          & x          & x          & x          & x \\
QuAC \cite{quac}                     & \checkmark & x          & x          & x & x          & x          & x          & \checkmark          & x \\
DailyDialog \cite{dailydialog}       & \checkmark & \checkmark & x          & x          & x          & x          & x          & x          & x \\
\hline\hline
\end{tabular}

\begin{center}
\footnotesize
\noindent\textbf{MT} — Multi-turn Conversations,  
\textbf{OD} — Open Domain Coverage,  
\textbf{PG} — Persona Grounding,  
\textbf{TC} — Topic Continuity,  
\textbf{KG} — Knowledge Graph Inclusion,  
\textbf{LT} — Long-term Memory Grounding, 
\textbf{ST} — Short-term Memory Grounding,  
\textbf{QA} — Question Answer Supervision,  
\textbf{AG} — Agent-based Generation.
\end{center}
\renewcommand{\arraystretch}{1.0}
\vspace{-4mm}
\end{table*}

\subsection{Post-processing Conversations}

The final post-processing stage prepares the dataset for distribution. The \textit{PostProcessing} module cleans and normalizes all generated records by removing duplicate entries, flattening nested structures, attaching source dataset conversations, and exporting each instance as JSON. Each processed record is represented as:

{\small
\begin{equation}
\mathcal{R} = (t_j,\ \mathcal{C}_j^{\text{final}},\ \mathcal{K}_j^{n},\ \mathcal{QA}_j,\ T_{source})
\end{equation}
}

\noindent where, \(t_j\) is the selected topic, \(\mathcal{C}_j^{\text{final}}\) is the finalized conversation, \(\mathcal{K}_j^{n}\) is the speaker-specific knowledge graphs for speaker \(n \in \{\textit{speaker1}, \textit{speaker2}\}\), \(\mathcal{QA}_j\) is the associated QA set, and $T_{source}$ denotes the total number of conversational turns in the source dataset for the speaker pair. This step ensures that the TGC dataset is standardized and ready for training or evaluation. 

\vspace{-1mm}
\section{TGC Dataset}
\vspace{-1mm}

We introduce TGC, a benchmark dataset for memory grounded fine-tuning and evaluation of LLMs in multi-turn, persona-grounded conversational settings. Unlike the session-based MSC dataset, TGC organizes conversations around high-level topics. Long-term memory is derived from knowledge graphs while short-term memory is derived from topic-guided conversations. Each persona pair includes up to top \(N\) topics, speaker personas with speaker-specific knowledge graphs, \(T\) conversational turns, and \(N_j\) memory-grounded QA pairs per topic. A sample structure of the TGC dataset is delineated in Figure~\ref{fig:dataset} in Appendix. %supplementary material.
%michael: please never hard code. Use proper referencing. 

%\vspace{0.4em}

\noindent\textbf{Distinctive Features --} Table~\ref{tab:dataset_comparison} highlights key differences between TGC and prior conversational datasets. TGC provides topic-guided conversations, speaker-specific knowledge graphs, fully automated generation without human annotation, and explicit short- and long-term context for memory evaluation in LLMs. %These characteristics make TGC a strong benchmark for memory-grounded QA and topic-guided conversation generation.

%\vspace{0.4em}

\noindent\textbf{Dataset Scale --} Table~\ref{tab:dataset_stats} provides an overview of the size and structural components of the TGC dataset, including personas, conversations, topics, knowledge graphs, and QA pairs. 

\renewcommand{\arraystretch}{1.01} 
\begin{table}[h]
\footnotesize
\centering
\caption{Scale of the TGC dataset.}
\label{tab:dataset_stats}
\vspace{-2mm}
\begin{tabular}{
    >{\centering\arraybackslash}m{5cm}|
    >{\centering\arraybackslash}m{2cm}
}
\hline\hline
\textbf{Dataset Records} & \textbf{Scale} \\
\hline
\# Persona Pairs & 1,001 \\

\hline
\# Conversational Turns & 88,500 \\
\hline
Avg. Turns / Topic & 30 \\
\hline
\# Topics & 2,950 \\
\hline
Avg. Topics / Persona Pair & 3 \\
\hline
\# Unique Tokens & 17,057 \\
\hline
\# Avg. Knowledge Graphs / Topic & 10 \\
\hline
\# Memory QA Pairs & 59,000 \\
\hline
QA Pairs / Topic & 20 \\
\hline\hline
\end{tabular}
\renewcommand{\arraystretch}{1.0} % Reset after this table
\vspace{-2mm}
\end{table}

\renewcommand{\arraystretch}{1.1} % Tighten row height
\begin{table}[h]
\footnotesize
\centering
\caption{Human evaluation of the dataset quality.}
\label{tab:human_eval_part1}
\vspace{-2mm}
\begin{tabular}{>{\centering\arraybackslash}p{1.9cm}|
                >{\centering\arraybackslash}p{2.5cm}|
                >{\centering\arraybackslash}p{0.8cm}|
                >{\centering\arraybackslash}p{0.8cm}}
\hline
\hline
\textbf{Type} & 
\textbf{Metric} & 
\textbf{MSC} & 
\textbf{TGC} \\
\hline

\multirow{5}{*}{\makecell[c]{Conversational\\Turns}} 
& Topic Consistency     & 64.12\% & \textbf{97.82\%} \\
& Factual Correctness   & \textbf{95.54\%} & 94.37\% \\
& Completeness          & 65.31\% & \textbf{88.22\%} \\
& Coherence             & 87.12\% & \textbf{94.67\%} \\
& Naturalness           & \textbf{98.76\%} & 97.11\% \\
\hline

\multirow{3}{*}{\makecell[c]{Knowledge\\Graphs}} 
& Topic Consistency     & --      & \textbf{90.36\%} \\
& Factual Correctness   & --      & \textbf{92.18\%} \\
& Relevance             & --      & \textbf{90.02\%} \\
\hline

\multirow{2}{*}{\makecell[c]{QA Pairs}} 
& Factual Correctness   & --      & \textbf{86.82\%} \\
& Relevance             & --      & \textbf{87.41\%} \\

\hline
\hline
\end{tabular}
\vspace{-5mm}
\end{table}

\renewcommand{\arraystretch}{1.23} 
\begin{table*}[h]
\footnotesize
\centering
\caption{Automatic evaluation of generated datasets across metrics.}
\label{tab:automatic_eval}
\vspace{-2mm}
\resizebox{0.98\linewidth}{!}
{
\begin{tabular}{
    >{\centering\arraybackslash}m{2.3cm}|
    >{\centering\arraybackslash}m{2.7cm}|
    >{\centering\arraybackslash}m{0.7cm}|
    >{\centering\arraybackslash}m{0.7cm}|
    >{\centering\arraybackslash}m{0.7cm}|
    >{\centering\arraybackslash}m{0.7cm}|
    >{\centering\arraybackslash}m{0.7cm}|
    >{\centering\arraybackslash}m{0.7cm}|
    >{\centering\arraybackslash}m{0.7cm}|
    >{\centering\arraybackslash}m{0.7cm}|
    >{\centering\arraybackslash}m{0.7cm}
}
\hline\hline
\textbf{Metric} & \textbf{Submetric} 
& \multicolumn{3}{c|}{\textbf{MSC Dataset}} 
& \multicolumn{3}{c|}{\textbf{PC Dataset}} 
& \multicolumn{3}{c}{\textbf{DD Dataset}} \\
\cline{3-11}
& & \textbf{TGC w/ v} & \textbf{TGC w/o v} & \textbf{MSC} 
  & \textbf{TGC w/ v} & \textbf{TGC w/o v} & \textbf{PC}
  & \textbf{TGC w/ v} & \textbf{TGC w/o v} & \textbf{DD} \\
\hline

\multirow{3}{*}{\begin{tabular}[c]{@{}c@{}}Conversational\\Flow (BERTScore) \end{tabular}}
& Precision   & \textbf{30.13} & 24.80 & 12.20 & 28.11 & 22.92 & 11.64 & 27.26 & 21.80 & 9.33 \\
& Recall    & \textbf{30.60} & 24.60 & 12.70 & 28.50 & 23.43 & 12.01 & 27.78 & 22.19 & 10.10 \\
& F1    & \textbf{30.36} & 24.70 & 12.44 & 28.30 & 23.17 & 11.82 & 27.52 & 21.99 & 9.70 \\
\hline

\multirow{2}{*}{\begin{tabular}[c]{@{}c@{}}Topic Consistency \end{tabular}}
& Self-BLEU & \textbf{50.25} & 43.23 & 40.50 & 49.80 & 44.80 & 38.70 & 49.10 & 45.50 & 39.01  \\
& Perplexity [$\geq$ 1] & \textbf{47.58} & 46.25 & 41.57 & 46.00 & 44.50 & 40.00 & 43.50 & 42.00 & 38.00 \\
\hline

\multirow{3}{*}{\begin{tabular}[c]{@{}c@{}}Content\\Similarity\end{tabular}}
& ROUGE-1  & \textbf{24.24} & 16.15 & 13.95 & 23.00 & 14.50 & 11.50 & 21.00 & 14.00 & 10.00 \\
& ROUGE-2  & \textbf{4.71} & 2.29 & 2.04 & 4.20 & 2.05 & 1.80 & 4.00 & 1.60 & 1.50 \\
& ROUGE-L  & \textbf{17.11} & 12.42 & 11.37 & 16.50 & 11.50 & 10.50 & 13.50 & 9.80 & 9.20 \\
\hline

\multirow{2}{*}{\begin{tabular}[c]{@{}c@{}}Semantic\\Understanding\end{tabular}}
& Semantic Similarity  & \textbf{13.36} & 6.85 & 6.26 & 12.00 & 6.00 & 5.80 & 12.50 & 6.10 & 5.88 \\
%& Topic Consistency    & 10.91 & 5.23 & 4.28 & 9.80 & 4.80 & 3.90 & 7.20 & 3.50 & 3.00 \\
& Entity Overlap       & \textbf{16.40} & 8.04 & 5.98 & 15.00 & 7.20 & 5.20 & 15.10 & 7.80 & 5.47 \\
\hline

\multirow{1}{*}{Readability}
& Flesch Reading Ease  & 85.91 & 73.85 & \textbf{87.92} & 82.00 & 72.00 & 84.50 & 78.00 & 68.00 & 80.00 \\
\hline\hline
\end{tabular}
}
\renewcommand{\arraystretch}{1.0}
\vspace{-3mm}
\end{table*}

\section{Experimental Results}
\vspace{-1mm}

\subsection{Human Evaluation of the Dataset Quality}
\label{sec:hu}
\vspace{-1mm}
We conducted a structured human evaluation comparing conversation quality in the TGC dataset against the MSC dataset. A 10\% sample of the said datasets was annotated by four NLP researchers. Each sampled conversation was rated across topical relevance, factual correctness, completeness, coherence, and naturalness using a 0–100 scale with scores averaged across annotators to reflect overall conversational quality. Since MSC does not contain knowledge graphs or QA pairs, topic consistency, factual correctness, and relevance were assessed only for the knowledge graph and QA components of the TGC.

As shown in Table~\ref{tab:human_eval_part1}, TGC achieved higher scores for conversational turns in most metrics. Topic consistency improved by +33.7 points, completeness by +22.9, and coherence by +7.5. Factual correctness and naturalness remained high for both datasets with MSC slightly higher due to its human-authored nature while TGC still achieved 94.37\% and 97.11\%, respectively. Knowledge graph evaluation further showed strong quality with TGC scoring 90.36\% in topic consistency, 92.18\% in factual correctness, and 90.02\% in relevance, indicating accurate and topic-aligned knowledge graph construction. TGC QA pairs were also evaluated as highly reliable, scoring 86.82\% in factual correctness and 87.41\% in relevance, demonstrating effective memory-grounded QA generation.

\vspace{-1mm}
\subsection{Ablation Evaluation on Dataset Quality}
As delineated in Table~\ref{tab:automatic_eval}, automatic evaluations were conducted across five metric categories on 10\% uniformly sampled subsets of three datasets: Multi-session Chat (MSC), PersonaChat (PC), and DailyDialog (DD). Although the final TGC dataset is derived from MSC, the additional two datasets were incorporated to assess the robustness and generalizability of the AgenticAI-DialogGen framework. For each source dataset, both validated (w/v) and unvalidated (w/o v) TGC variants were generated and compared against the source dataset to evaluate the contribution of the \textit{ConversationValidator} agent.

\vspace{1mm}
\noindent\textbf{Conversational Flow --} TGC w/v generated from MSC achieved the highest BERTScore F1 across all datasets showing the strongest gain (30.36). TGC w/v from PersonaChat and DailyDialog followed the same trend, indicating clearer and more coherent turn transitions.

\renewcommand{\arraystretch}{1} % Reduce row height
\FloatBarrier
\begin{table*}[!tb]
\footnotesize % Slightly smaller than \small
\centering
\caption{Performance comparison of LLMs on memory QA evaluation (zero-shot and few-shot settings).}
\label{tab:model_comparison}
\resizebox{0.97\linewidth}{!}
{
\begin{tabular}{
    >{\centering\arraybackslash}m{3.2cm}|
    >{\centering\arraybackslash}m{2.8cm}|
    >{\centering\arraybackslash}m{1.8cm}|
    >{\centering\arraybackslash}m{1.8cm}|
    >{\centering\arraybackslash}m{1.8cm}|
    >{\centering\arraybackslash}m{2cm}
}
\hline\hline
\textbf{Model} & \textbf{Evaluated Dataset} & \textbf{Precision} & \textbf{Recall} & \textbf{F1} & \textbf{Exact Match} \\
\hline
\multicolumn{6}{c}{\textbf{Zero-shot Evaluation}} \\
\hline
\multirow{3}{*}{\begin{tabular}{c}GPT-4\\\cite{GPT4Tec}\end{tabular}} 
    & TGC / KG      & \textbf{82.10} & \textbf{85.50} & \textbf{83.77} & \textbf{35.87} \\
    & TGC w/o KG    & 69.81 & 73.43 & 71.57 & 26.74 \\
    & MSC           & 60.23 & 63.94 & 62.03 & 20.95 \\
\hline
\multirow{3}{*}{\begin{tabular}{c}Gemini 2.5 Flash\\\cite{Gemini2.5}\end{tabular}}
    & TGC / KG      & \textbf{80.11} & \textbf{84.23} & \textbf{82.12} & \textbf{35.21} \\
    & TGC w/o KG    & 67.52 & 72.30 & 69.83 & 22.31 \\
    & MSC           & 58.67 & 60.16 & 59.41 & 17.87 \\
\hline
\multirow{3}{*}{\begin{tabular}{c}Claude 3.5 Sonnet\\\cite{Claude}\end{tabular}}
    & TGC / KG      & \textbf{83.24} & \textbf{86.11} & \textbf{84.65} & \textbf{39.49} \\
    & TGC w/o KG    & 71.66 & 73.98 & 72.80 & 25.31 \\
    & MSC           & 61.42 & 62.34 & 61.88 & 17.52 \\
\hline
\multirow{3}{*}{\begin{tabular}{c}DeepSeek R1\\\cite{deepseek}\end{tabular}}
    & TGC / KG      & \textbf{78.58} & \textbf{82.25} & \textbf{80.37} & \textbf{33.01} \\
    & TGC w/o KG    & 66.17 & 69.46 & 67.78 & 26.40 \\
    & MSC           & 55.71 & 58.82 & 57.22 & 19.21 \\
\hline
\multicolumn{6}{c}{\textbf{Fine-tune Evaluation}} \\
\hline
\multirow{3}{*}{\begin{tabular}{c}mistral-7b-v0.3-bnb-4bit\\\cite{mishral}\end{tabular}}
    & TGC / KG      & \textbf{85.51} & \textbf{89.29} & \textbf{87.36} & \textbf{38.38} \\
    & TGC w/o KG    & 72.23 & 75.61 & 73.88 & 28.05 \\
    & MSC           & 61.81 & 64.07 & 62.92 & 24.11 \\
\hline
\multirow{3}{*}{\begin{tabular}{c}Qwen2.5-7\\\cite{Qwen}\end{tabular}}
    & TGC / KG      & \textbf{81.53} & \textbf{85.02} & \textbf{83.24} & 35.81 \\
    & TGC w/o KG    & 69.57 & 72.88 & 71.18 & \textbf{31.51} \\
    & MSC           & 58.11 & 61.08 & 59.56 & 23.33 \\
\hline
\multirow{3}{*}{\begin{tabular}{c}Phi-3.5-mini-instruct\\\cite{phi3}\end{tabular}}
    & TGC / KG      & \textbf{79.21} & \textbf{83.43} & \textbf{81.26} & \textbf{34.29} \\
    & TGC w/o KG    & 67.28 & 71.06 & 69.12 & 29.05 \\
    & MSC           & 56.88 & 60.34 & 58.56 & 23.17 \\
\hline
\multirow{3}{*}{\begin{tabular}{c}Meta-Llama-3.1-8B\\\cite{llama3}\end{tabular}}
    & TGC / KG      & \textbf{77.55} & \textbf{80.63} & \textbf{79.06} & \textbf{32.01} \\
    & TGC w/o KG    & 65.88 & 69.19 & 67.49 & 25.02 \\
    & MSC           & 55.01 & 58.36 & 56.63 & 22.62 \\
\hline\hline
\end{tabular}
}
\vspace{-4mm}
\end{table*}

\vspace{1mm}
\noindent\textbf{Topic Consistency --} TGC w/v achieved higher Self-BLEU and perplexity across all datasets, indicating stronger topic consistency and controlled lexical variation. Since conversations in TGC are explicitly generated around fixed topics, higher Self-BLEU reflects reduced topical drift rather than undesirable repetition. At the same time, increased perplexity in TGC w/v generated from MSC (47.58), PersonaChat (46.00), and DailyDialog (43.50) suggests richer linguistic expression within topic-constrained conversations, supporting more natural long-context reasoning.

\vspace{1mm}
\noindent\textbf{Content Similarity --} ROUGE scores consistently favored TGC w/v generated from MSC with similar relative improvements observed for PersonaChat and DailyDialog, indicating improved lexical continuity and topic grounding.

\vspace{1mm}
\noindent\textbf{Semantic Understanding --} TGC w/v scored the highest in semantic similarity and entity overlap across all datasets. TGC w/v generated from MSC shows the largest increase, while TGC w/v from PersonaChat and DailyDialog exhibit the same pattern, confirming improved meaning preservation and referential clarity.

\vspace{1mm}
\noindent\textbf{Readability --} Source datasets showed slightly higher Flesch Reading Ease scores because of the human authored nature while TGC w/v still maintains strong readability across MSC (85.91), PersonaChat (82.00), and DailyDialog (78.00).

Overall, the results demonstrate the robustness of the AgenticAI-DialogGen framework and how the \textit{ConversationValidator} enhances quality across diverse datasets. Additionally, detailed definitions of the automatic evaluation metrics used in Table~\ref{tab:automatic_eval} are provided in Section \ref{sec:metrics} in Appendix. 
%the supplementary material.

\vspace{-1mm}
\subsection{Ablation Evaluation of LLMs on Memory Grounded QA Tasks}

To assess the TGC dataset for memory grounded QA, we evaluated 8 LLMs on a task probing both short- and long-term recall. QA pairs for TGC and MSC were generated using same LLM model to ensure consistency. These pairs simulated realistic memory checks where \textit{speaker 1} asks about its prior events or preferences and \textit{speaker 2} provides brief factual responses. This allows evaluation of an LLM’s ability to retrieve speaker and topic-specific information grounded in structured short- and long-term context.

Table~\ref{tab:model_comparison} presents results across precision, recall, F1, and exact match. We compare three settings: (1) TGC with knowledge graphs (TGC / KG), (2) TGC without knowledge graphs (TGC w/o KG), and (3) MSC, by isolating the role of knowledge graphs. TGC / KG combines topic-guided turns with speaker-specific knowledge graphs while TGC w/o KG combines topic-guided turns with raw MSC conversations. From 59,000 QA pairs, we used an 85/5/10 for train/validation/test split at the persona pair level. TGC / KG consistently produced the highest scores. GPT-4 reached an F1 of 83.77 with TGC / KG, dropping to 71.57 without KG and 62.03 on MSC, confirming the importance of structured memory. Among zero-shot models, GPT-4 and Claude 3.5 Sonnet performed the strongest, while DeepSeek R1 showed lower exact match scores. Fine-tuned lightweight models, including Mistral-7B, Qwen2.5, Phi-3.5, and Meta-LLaMA-3.1, outperformed larger zero-shot models and Mistral-7B achieved the highest F1 of 87.36, demonstrating the advantages of training on structured inputs, i.e., knowledge graphs. The 10–15 point gap between TGC / KG and TGC w/o KG emphasized the value of explicit memory encoding.  The results further demonstrate that lightweight models can display strong memory-aware behavior when fine-tuned on TGC, validating its design as both a benchmark and a training resource for memory-grounded conversational tasks.

\vspace{-1mm}
\section{Conclusion and Future Work}
\vspace{-1mm}
In this paper, we introduced AgenticAI-DialogGen, a modular agent-based framework for generating persona-grounded and topic-guided conversations without human supervision. Built on the Multi-Session Chat (MSC) dataset, the framework converts session-based conversations into structured, topic-guided interactions through coordinated LLM modules and agents. We also release the TGC dataset which provides short-term context and long-term context in two formats: (1) token-efficient knowledge graphs that serve as structured memory and (2) MSC conversations that serve as unstructured memory. This dual-memory design supports a broad range of memory-grounded tasks. Human and automatic evaluations show that TGC improves discourse quality, strengthens memory-grounded QA performance, and enables lightweight LLMs to exhibit strong memory-aware behavior, often surpassing larger zero-shot models. 

Future work will explore learning-based strategies, e.g., reinforcement learning, to provide targeted feedback signals that improve the accuracy and robustness of conversational knowledge graph construction, thereby supporting stronger 
long-context reasoning.

\section{Limitations}

AgenticAI-DialogGen relies on LLMs for knowledge extraction, topic extraction, persona generation, and conversational simulation. As a result, factual inaccuracies, biases, or reasoning errors inherent to the underlying LLMs may propagate into the generated knowledge graphs, personas, and conversations despite the inclusion of validation and refinement stages. In addition, although AgenticAI-DialogGen is fully automated and does not require human annotation, dataset generation requires access to LLM APIs and associated computational resources. Finally, the characteristics of the generated conversations may be influenced by design choices, i.e., prompt formulations, agent configurations, and validation thresholds, which are provided as configurable inputs in the framework.

\section{Ethical Considerations}
This work used only publicly available datasets, i.e., MSC, PersonaChat, and DailyDialog, together with synthetic content generated by LLM agents. No real individuals are represented in the generated personas, knowledge graphs, or conversations, and all speaker references are abstract and non-identifying. Although LLMs may inherit biases from pretrained models, validation and refinement steps are applied to mitigate inappropriate or off-topic content. The TGC dataset is intended solely for research on short- and long-context conversational modeling and should not be deployed in sensitive or decision-making settings. All source datasets were used in compliance with their licensing conditions and intended research purposes.

%\section{Acknowledgments}

%\balance
%\bibliographystyle{plain}
\bibliography{ReferencesTechnical}

\clearpage
\appendix

%michael: it is not right for the below title. Please give a different section name rather than Appendix. Just fixed the problem. 
%\section{Supplementary Material}
%\label{sec:appendix}

\section{TGC Dataset Structure}

\begin{figure}[!tb]
  \centering
  \includegraphics[width=1\linewidth]{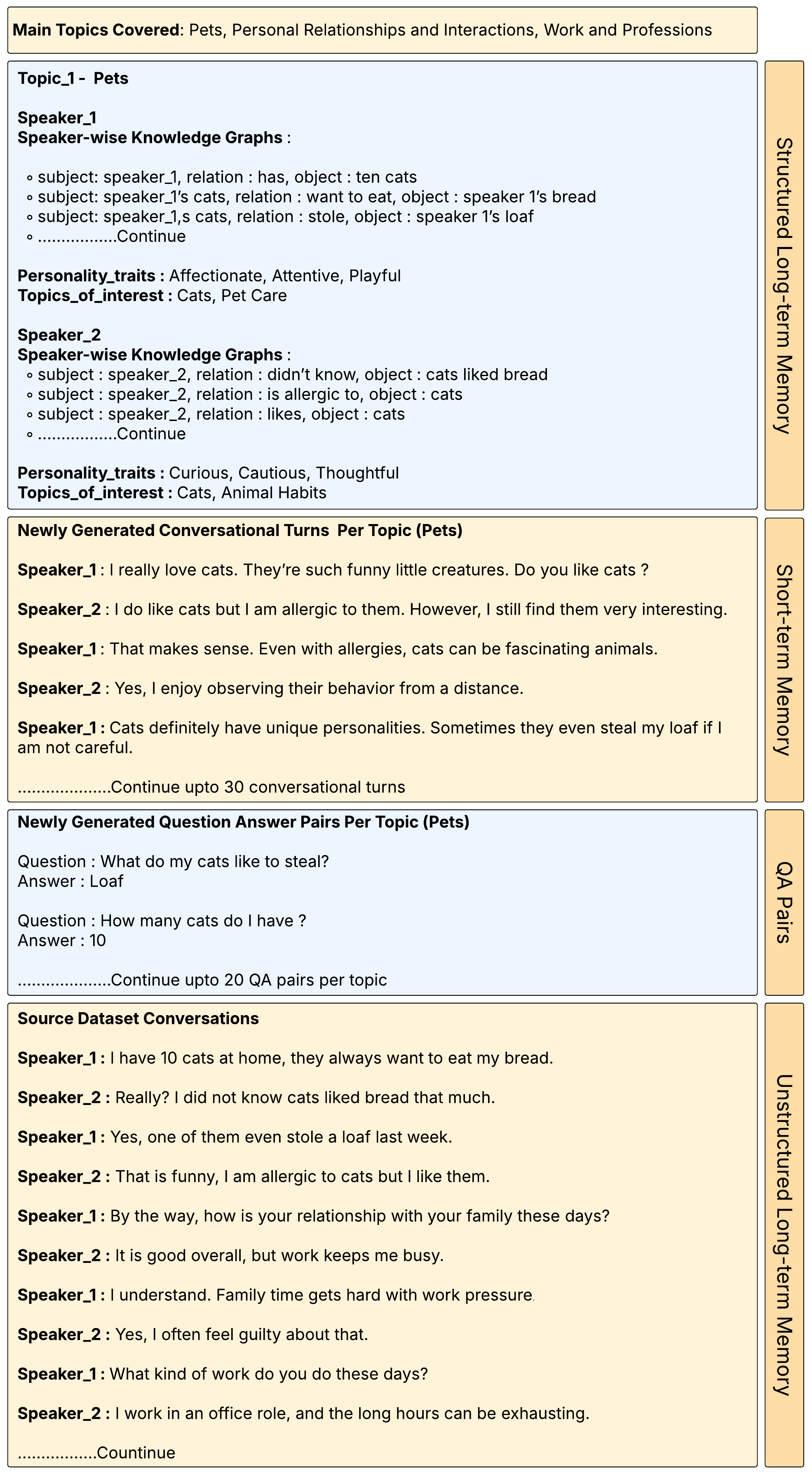}
   % \captionsetup{justification=centering}

  \caption{Overview of a sample TGC dataset for a single topic.}
  \label{fig:dataset}
  \vspace{-3mm}
  \end{figure}

Figure \ref{fig:dataset} illustrates the structure of a sample TGC instance for a given speaker pair and selected topic. For each topic, the dataset explicitly separates structured long-term memory, unstructured long-term memory, and short-term memory. Structured long-term memory is represented through speaker-specific knowledge graphs, along with derived personality traits and topics of interest, which is constructed from MSC conversations. Unstructured long-term memory consists of the original MSC dataset conversations associated with the same speaker-pair. Short-term memory comprises newly generated topic-guided conversational turns simulated for the selected topic using the corresponding speaker personas. In addition, each topic includes memory-grounded QA pairs generated from speaker-specific knowledge graphs and newly generated conversational turns, where \textit{speaker1} asks questions about its prior events or preferences and \textit{speaker2} provides brief factual responses. This structure supports controlled evaluation of topic continuity, persona consistency, and memory-grounded reasoning within a unified topic-centric setting.

\section{Distributional Analysis of Topic Consistency and Conversational Flow}

The topic consistency and conversational flow metrics from Table~\ref{tab:automatic_eval} are further examined through their distribution across individual conversations. In the MSC dataset, each conversation corresponds to a complete session between a persona pair, whereas in the TGC dataset, each conversation corresponds to a single topic-specific conversation for a persona pair. Topic consistency is measured at the conversation level using Self-BLEU which captures the degree of lexical overlap across turns within a conversation. Conversational flow is measured as the session-level BERTScore F1 between consecutive turns, reflecting turn-to-turn semantic alignment. This distributional analysis enables a finer-grained comparison of conversational quality beyond aggregate averages by revealing variability across individual conversations.

\begin{figure*}[tbp]
\small
  \centering
  \begin{minipage}[t]{0.48\linewidth}
    \centering
    \includegraphics[width=0.99\linewidth]{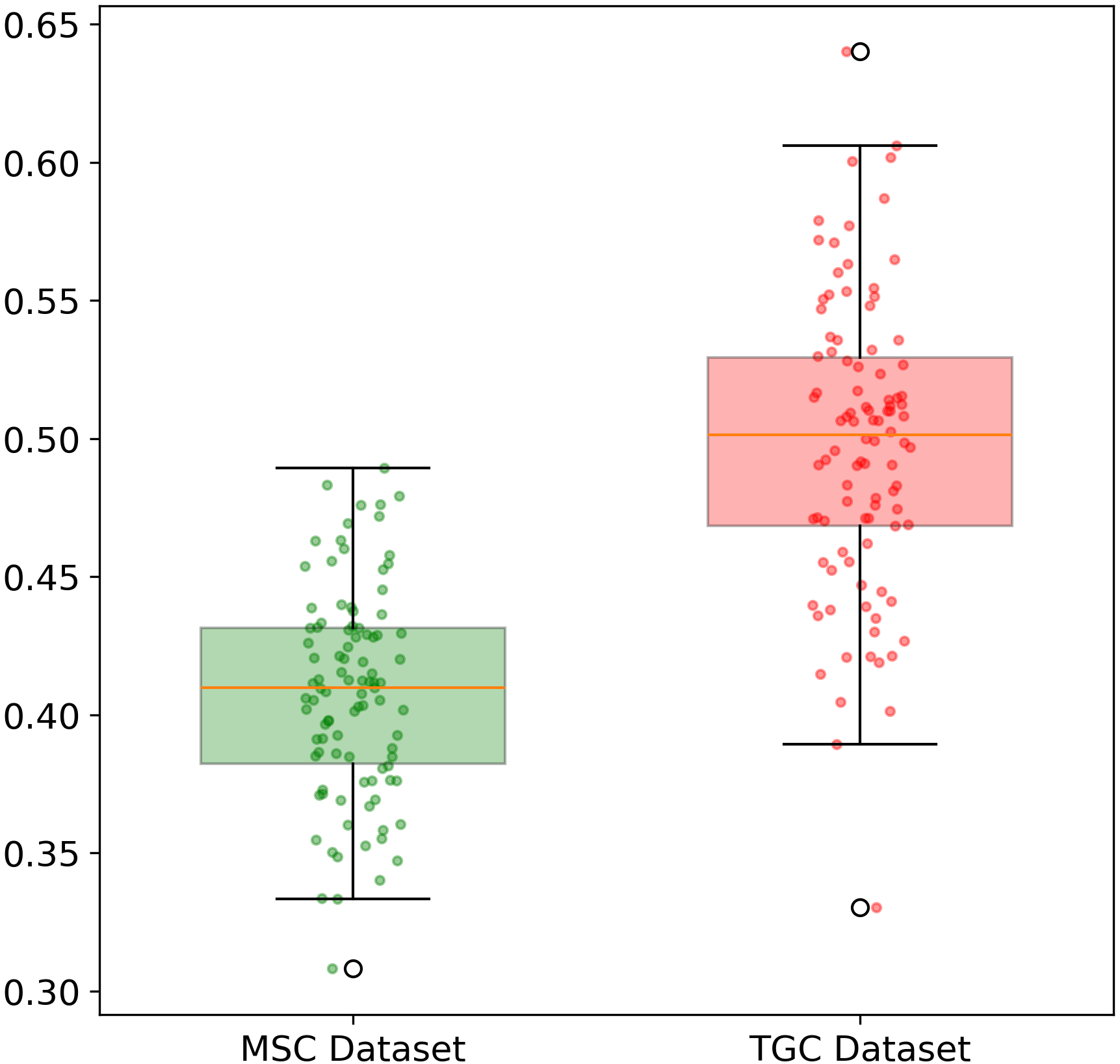}
    \caption{\small Topic consistency between MSC and TGC datasets.}
    \label{fig:topic_consistency}

  \end{minipage}
  \hfill
  \begin{minipage}[t]{0.48\linewidth}
    \centering
    \includegraphics[width=0.99\linewidth]{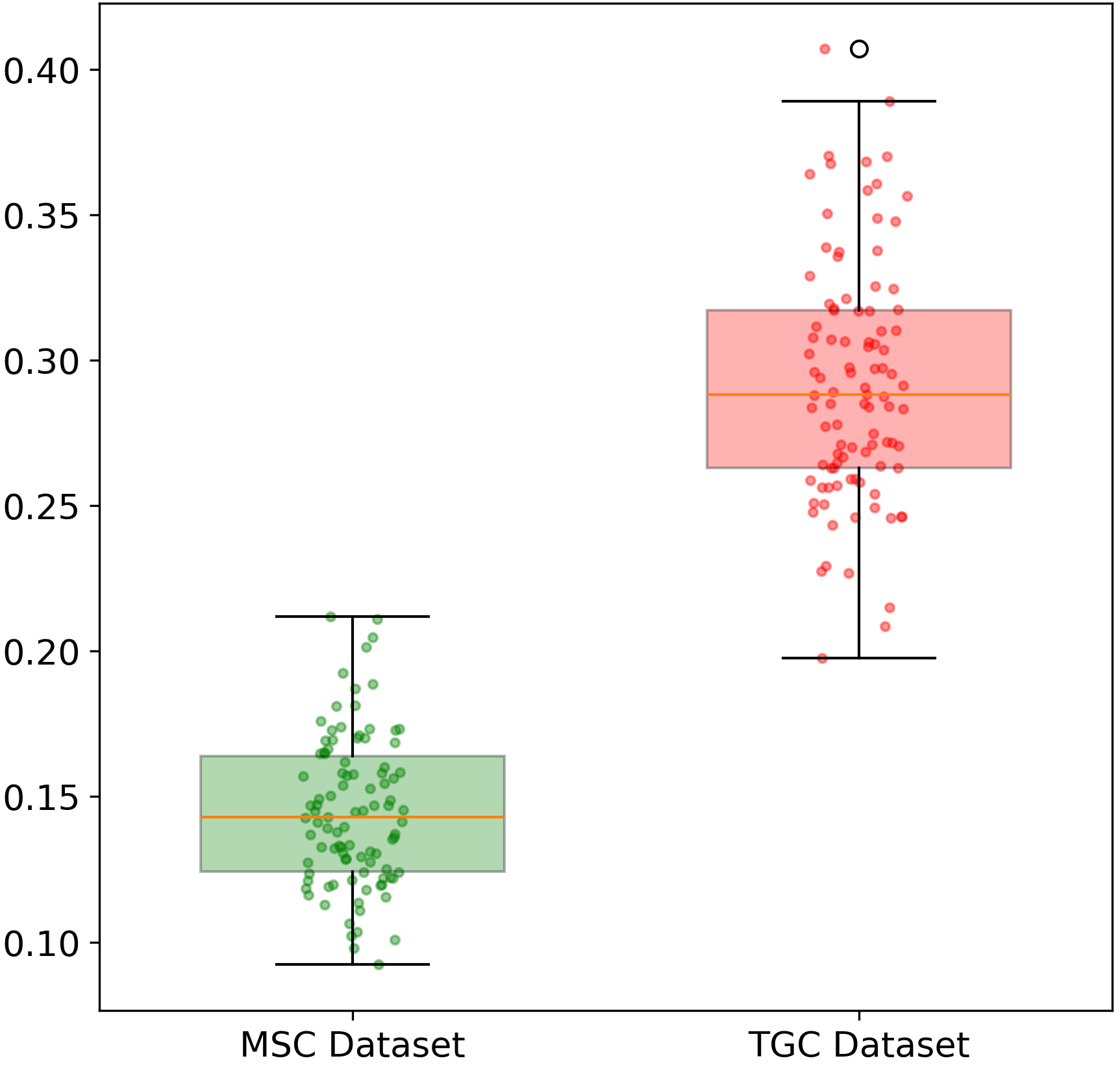}
    \caption{\small Conversational flow between MSC and TGC datasets.}
    \label{fig:conversational_flow}
  \end{minipage}
%  \vspace{-4mm}
\end{figure*}

Figure~\ref{fig:topic_consistency} shows that the TGC dataset exhibits substantially higher topic consistency than the MSC dataset, with a median session Self-BLEU of approximately 0.50 for TGC compared to 0.41 for MSC. Figure~\ref{fig:conversational_flow} similarly demonstrates stronger conversational flow in the TGC dataset, with a median session-level BERTScore F1 of approximately 0.29 for TGC versus 0.14 for MSC, indicating smoother semantic transitions between turns in topic-guided conversations. These distributional results reinforce the aggregate findings that topic-guided simulation improves both topical adherence and contextual alignment in multi-turn conversations.

\section{Implementation Details}
The complete source code for the AgenticAI-DialogGen framework, together with the full TGC dataset, will be publicly released upon paper acceptance. A preview subset of the TGC dataset, including topic-guided conversations, speaker-specific knowledge graphs, and memory-grounded QA pairs, is currently available at:
\url{https://github.com/ConversatAI/AgenticAI-DialogGen.git}. 

In the current instantiation of the TGC dataset used in this work, we set the maximum number of extracted topics per persona pair to \(N = 3\). For each topic, the framework generates \(T = 30\) topic-guided conversational turns and \(N_j = 20\) memory-grounded question answering pairs. These parameter settings were chosen to balance conversational depth, topic coverage, and computational efficiency, and can be adjusted easily through configuration files in the released implementation.

The Agentic-DialogGen framework is implemented in Python and follows a modular and agent-based design where each stage is encapsulated as a standalone module or agent. These components include preprocessing, knowledge extraction, topic extraction, topic-specific knowledge graph construction, persona generation, agent-based conversation simulation, validation, refinement, and QA pair generation. All agents interact through structured inputs and outputs with predefined schemas to ensure consistency across the framework. All prompts used by the LLM agents are stored externally and invoked programmatically, enabling controlled prompt reuse and facilitating future extensions or ablation studies. Prompt templates would be released alongside the source code to support procedural reproducibility. 

\subsection{Threshold Calibration}

The validation agent assigns two continuous scores in the range [0, 1] to each simulated conversation, namely a topical adherence score and an overall quality score. A conversation is accepted only if both scores exceed user-defined threshold values specified in the framework configuration.

In our experiments, we set both thresholds to 0.85. This value was calibrated using a held-out development pool of generated conversations by inspecting the empirical score distributions and manually reviewing conversations around different cutoff values. Conversations with scores below 0.8 frequently exhibited topical drift, repetition, or unnatural phrasing, whereas conversations scoring above 0.85 were consistently judged to be on topic, coherent, and linguistically natural. Selecting a threshold of 0.85 therefore removed the majority of low-quality generations while retaining a sufficient number of high-quality instances for dataset construction. Allowing threshold values to be configured enables users of the AgenticAI-DialogGen framework to trade off generation quality against data volume based on their application requirements. This design choice aligns with confidence-based data selection practices commonly adopted in language model training frameworks, where high-confidence thresholds are used to filter synthetic or weakly supervised data. 

\subsection{Language Model Configuration}
The AgenticAI-DialogGen framework is model-independent and is designed to support interchangeable LLM through a configurable interface. All LLM-specific settings, including the model identifier, API endpoint, decoding parameters, and rate limits, are specified in a central configuration file. This design allows users of the framework to seamlessly substitute different proprietary or open-source LLMs without modifying the core framework logic. 

In our experiments, all agents and modules within the AgenticAI-DialogGen framework, including preprocessing, knowledge extraction, topic extraction, topic-specific knowledge graph construction, persona generation, agent-based conversation simulation, validation, refinement, and QA pair generation, were instantiated using the same underlying LLM to ensure consistency across stages. Specifically, we used GPT-4o as the backbone model in our AgenticAI-DialogGen framework. GPT-4o is explicitly optimized for natural, real-time, multi-turn conversational interaction, with training and evaluation focused on dialogue coherence and human-like responses \cite{4o,DialogueForge}. Since our objective was to synthesize realistic topic-guided conversations rather than optimize for complex reasoning, GPT-4o was proved well suited for our framework.

\section{Automatic Evaluation Metrics}
\label{sec:metrics}

This section briefly explains the automatic evaluation metrics reported in Table~\ref{tab:automatic_eval} and clarifies their usage in the context of topic-guided conversational dataset evaluation.

\subsection{Conversational Flow (BERTScore)}
%\vspace{-1mm}
Conversational flow is measured using BERTScore computed between consecutive conversational turns, capturing semantic alignment and coherence in turn-to-turn transitions using contextual embeddings from bert-base-uncased model. 

\subsection{Topic Consistency (Self-BLEU and Perplexity)}
%\vspace{-1mm}
Topic consistency is evaluated using Self-BLEU which measures n-gram overlap across turns within a conversation, and perplexity is computed using an n-gram language model to assess predictability and lexical variation. While Self-BLEU is often used as a diversity metric where lower values indicate greater variation, in our topic-guided setting higher Self-BLEU reflects reduced topical drift across turns. Perplexity is used comparatively as a proxy for controlled linguistic variability within topic-constrained conversations rather than as an absolute fluency measure.

%michael: there is no space limit. So there is no need to cut space. 
\subsection{Content Similarity (ROUGE)}
%\vspace{-1mm}
Content similarity is assessed using ROUGE-1, ROUGE-2, and ROUGE-L, which quantify lexical overlap and longest common subsequence between consecutive turns. These metrics are employed to capture local grounding and continuity between adjacent responses rather than semantic equivalence at the conversation level.

\subsection{Semantic Understanding (Semantic Similarity and Entity Overlap)}
%\vspace{-1mm}
Semantic similarity is measured via vector-based similarity between consecutive turns, capturing alignment in meaning across responses. Entity overlap tracks the recurrence of content terms across turns, reflecting referential consistency and preservation of conversational context.
Together, these measures provide complementary signals of semantic continuity beyond surface-level lexical overlap.

\subsection{Readability (Flesch Reading Ease)}
%\vspace{-1mm}
Readability is evaluated using the Flesch Reading Ease score which estimates textual accessibility based on sentence length and word complexity. This metric is used descriptively to compare the structural readability of generated and source conversations.

\end{document}